%% file: main.tex
\documentclass{article} % For LaTeX2e
\usepackage{iclr2025_conference,times}
\usepackage{graphicx}
\usepackage{amsmath}
\usepackage[ruled,vlined]{algorithm2e}
\usepackage{algorithmic}
\usepackage{booktabs}
\usepackage{xcolor}
\usepackage{pifont}
\usepackage{comment}
\usepackage{multirow}
\usepackage{multicol}

\input{math_commands.tex}

\usepackage{hyperref}
\usepackage{url}
\usepackage{subcaption}
\usepackage[capitalize]{cleveref}

\definecolor{yes}{RGB}{51,160,44}
\definecolor{no}{RGB}{228,26,28}

\newcommand{\inputs}{\ensuremath{\mathcal{I}}}
\newcommand{\targets}{\ensuremath{\mathcal{T}}}
\newcommand{\light}[1]{\textcolor{gray}{#1}}

\newcommand{\auxiliary}{\ensuremath{\mathcal{A}}}
\newcommand{\base}{\ensuremath{\mathcal{B}}}
\newcommand{\predictor}{\ensuremath{\mathcal{P}}}
\newcommand{\auxiliarykv}{\ensuremath{\mathcal{KV}_\mathcal{A}}}
\newcommand{\basekv}{\ensuremath{\mathcal{KV}_\mathcal{B}}}
\newcommand{\predictedbasekv}{\ensuremath{\hat{\mathcal{KV_\mathcal{B}}}}}
\newcommand{\logitsa}{\ensuremath{\mathcal{A}(\mathcal{I})}}
\newcommand{\logitsb}{\ensuremath{\mathcal{B}(\mathcal{I})}}
\newcommand{\baseloss}{\ensuremath{\mathcal{L}_B}}
\newcommand{\auxiliaryloss}{\ensuremath{\mathcal{L}_A}}
\newcommand{\consistencyloss}{\ensuremath{\mathcal{L}_C}}

\title{KV Prediction for Improved Time to First \\ Token}

\author{Maxwell Horton\thanks{Correspondence to \texttt{mchorton@apple.com}.} , Qingqing Cao, Chenfan Sun, Yanzi Jin, Sachin Mehta\thanks{Work done at Apple. Now at Meta.} , Mohammad Rastegari$^\dagger$, \& \and \textbf{Moin Nabi} \\Apple
}

\iclrfinalcopy % Uncomment for camera-ready version, but NOT for submission.
\arxivcopy
\begin{document}

\maketitle

\begin{abstract} \label{sec:abstract}
Inference with transformer-based language models begins with a prompt processing step. In this step, the model generates the first output token and stores the KV cache needed for future generation steps. This prompt processing step can be computationally expensive, taking 10s of seconds or more for billion-parameter models on edge devices when prompt lengths or batch sizes rise. This degrades user experience by introducing significant latency into the model's outputs. To reduce the time spent producing the first output (known as the ``time to first token'', or \textit{TTFT}) of a pretrained model, we introduce a novel method called KV Prediction. In our method, a small \textit{auxiliary} model is used to process the prompt and produce an approximation of the KV cache used by a \textit{base} model. This approximated KV cache is then used with the base model for autoregressive generation without the need to query the auxiliary model again. We demonstrate that our method produces a pareto-optimal efficiency-accuracy trade-off when compared to baselines. On TriviaQA, we demonstrate relative accuracy improvements in the range of $15\%-50\%$ across a range of TTFT FLOPs budgets. We also demonstrate accuracy improvements of up to $30\%$ on HumanEval python code completion at fixed TTFT FLOPs budgets. Additionally, we benchmark models on an Apple M2 Pro CPU and demonstrate that our improvement in FLOPs translates to a TTFT speedup on hardware. We release our code at \url{https://github.com/apple/corenet/tree/main/projects/kv-prediction}.

%We evaluate on question answering and code completion tasks, demonstrating a substantial improvement over baselines at iso-FLOP computation. We also analyze the runtime characteristics of our method on-device.
\end{abstract}

\section{Introduction} \label{sec:introduction}
% Paragraph about LLM efficiency being important.
Large language models (LLMs) have demonstrated impressive capabilities on many downstream tasks \citep{Gunter2024AppleIF,Achiam2023GPT4TR,Chowdhery2022PaLMSL,Abdin2024Phi3TR}. However, the high computational cost of running large language models results in limited capabilities for on-device inference. On-device inference is essential for privacy, latency, energy efficiency, and performance in limited-connectivity areas \citep{Frantar2022GPTQAP,AlizadehVahid2023LLMIA,Stojkovic2024TowardsGL}. For these reasons, LLM efficiency remains an important and active area of research.

% Paragraph about prompt processing time being the bottleneck in many circumstances.
%\qc{we can discuss most work focus on reducing end to end latency or optimizing KV-cache, and motivate compelling use cases for TTFT.}
LLM inference with the popular transformer \citep{Vaswani2017AttentionIA} architecture consists of two phases. First, in the \emph{prompt processing} phase, the model processes an input prompt to populate the KV cache. Next, in the \emph{generation} phase, the model autoregressively generates output tokens. Many recent works focus on improving generation time through approximations of the KV cache \citep{Wu2024LayerCondensedKC,Jiang2023LLMLinguaCP,Ge2023ModelTY,Ge2023IncontextAF,Li2023CompressingCT}, but only a few recent works have explored improving the prompt processing time \citep{Fu2024LazyLLMDT}.  % As we will demonstrate in \cref{sec:motivation}, prompt processing time is significant in many real-world scenarios.

\begin{figure}
    \centering
    \includegraphics[width=\textwidth]{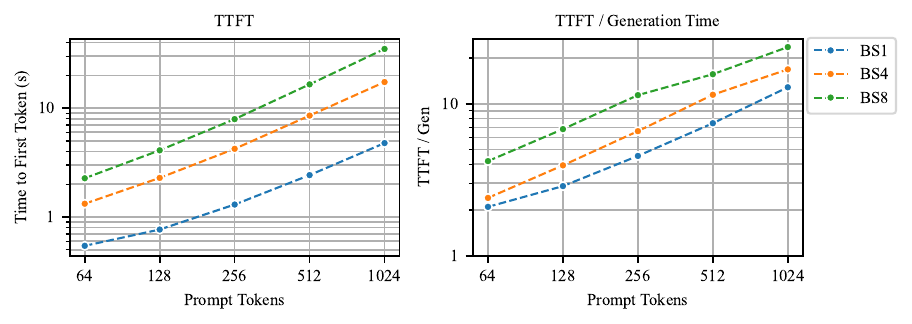}
    \caption{Time to First Token (TTFT) and ratio of TTFT to generation time for an OpenELM 3B model on an M2 Pro CPU \citep{everymac} with 32GB of RAM. We evaluate at batch sizes 1, 4, and 8.}
    \label{fig:timing-cpu}
\end{figure}

%\qc{should we discus existing solutions to reducing TTFT? like pruning, smaller models etc. we can say those techniques are orthogonal and we are better than directly using smaller models (or refer empirical results).  }
% One benefit of improving is TTFT being important.
Improving prompt processing time allows an application using the LLM to begin sending outputs to the user earlier. The ``time to first token'' (TTFT) refers to the length of time between a user's input query and the production of the first output token. In scenarios such as chatting with a user, the TTFT may be a more important runtime experience metric than autoregressive generation time, since the user can begin consuming outputs after the first token is produced. For on-device models, prompt processing times can be intolerably slow (up to 10s of seconds, \cref{fig:timing-cpu}). Reducing TTFT in these cases enables a better user experience.

% (Begin our contributions)
% We present a method to improve prompt processing time with minimal reduction of accuracy by using a smaller *auxiliary* network to predict the KV cache of the larger base network.
We present a method to improve TTFT by processing the prompt with a small \textit{auxiliary} transformer model. Our method runs the auxiliary model and stores its KV cache. It then uses a learned linear projection to predict the KV cache of another transformer model (the \textit{base} model) using only the KV cache of the auxiliary model as input. After the KV cache is predicted, inference continues exclusively with the base model. As a result, no runtime overhead is introduced during generation.

% We demonstrate that our method improves prompt processing time with minimal reduction of accuracy.
We demonstrate that our method improves over the efficiency-accuracy trade-off achievable by our baselines. For example, on TriviaQA \citep{Joshi2017TriviaQAAL}, our method improves accuracy retention by $15\%-50\%$ compared to baselines at equal TTFT FLOP counts. On HumanEval \citep{Chen2021EvaluatingLL} python code completion, we demonstrate accuracy improvements up to $30\%$ over baselines at equal TTFT FLOP counts. We also provide on-device timing experiments to demonstrate that our FLOPs gains translate to on-device runtime improvements.

% Summary of contributions.
Our contributions are as follows: (1) We develop a novel method called KV Prediction for using a smaller auxiliary model to predict the KV cache of a larger base model. (2) We show that our model produces a stronger efficiency-accuracy trade-off compared to baselines. (3) We analyze the runtime characteristics of KV Prediction models on-device. (4) We release our code at \url{https://github.com/apple/corenet/tree/main/projects/kv-prediction}.

The rest of the paper is organized as follows. In \cref{sec:related-work}, we give an overview of related work. In \cref{sec:motivation}, we motivate our work, demonstrating that TTFT can be problematically high during on-device inference. In \cref{sec:method}, we describe our KV Prediction method. In \cref{sec:experimental-setup}, we describe our experimental setup for evaluating KV Prediction. In \cref{sec:results}, we give our main results, showing that our method obtains the pareto-optimal efficiency-accuracy trade-off. In \cref{sec:analysis}, we analyze our predicted KV caches. In \cref{sec:conclusion}, we conclude.

\section{Related Work} \label{sec:related-work}
%\qc{maybe we can group the paragraphs into fewer ones, like TTFT optimizations, Activation Compression (KV cache is also a kind of activation, or just representation compression).  }
\textbf{On-Device TTFT Efficiency:} Few works have explored our problem domain of improving on-device TTFT. LazyLLM \citep{Fu2024LazyLLMDT} uses an attention-based token dropping strategy to drop unneeded tokens at inference time. These tokens can be revived later during the generation phase. Their evaluations are on GPU, but their technique is applicable to on-device inference. Random token dropping \citep{Yao2022RandomLTDRA} and static token pruning are both studied in \cite{Fu2024LazyLLMDT} as methods for improving TTFT. These methods are our baselines.

\textbf{Server-Side TTFT Efficiency:} Cachegen \citep{Liu2023CacheGenFC} compresses KV caches for faster TTFT, but their setup assumes that a precomputed, pre-compressed KV cache is stored on a server that is available over network. Another similar server-based approach, CritiPrefill \citep{Lv2024CritiPrefillAS}, performs attention-based pruning of KV cache segments on a per-query basis. Our investigation differs from these in that we focus on on-device TTFT improvements.

\textbf{Context Compression:} Previous works have investigated compressing the KV cache for improving generation efficiency. PyramidKV \citep{Cai2024PyramidKVDK}, StreamingLLM \citep{Xiao2023EfficientSL}, SnapKV \citep{Li2024SnapKVLK}, and Model Tells You What To Discard \citep{Ge2023ModelTY} all compress the KV cache along the token dimension by observing attention scores and pruning irrelevant tokens. However, their methods compute a forward pass before pruning. Thus, they improve generation time, but not TTFT. Layer-Condensed KV Cache \citep{Wu2024LayerCondensedKC} develops a method for compressing an $N$-layer KV cache into a $1$-layer KV cache, but requires $9$ prompt processing steps. Thus, their method improves generation time and memory usage but negatively impacts TTFT.

A few recent works have explored compressing the input context using a separate network \citep{Jiang2023LLMLinguaCP,Jiang2023LongLLMLinguaAA,Ge2023IncontextAF,Li2023CompressingCT}. However, their purpose is to improve generation time, not TTFT. As shown in \cite{Fu2024LazyLLMDT}, such methods can increase TTFT due to more expensive prompt processing.

\textbf{Sharing Activations Between Models:} Similar to our method, Tandem Transformers \citep{AishwaryaP2024TandemTF} uses two different models that share hidden activations. However, their focus is on improving the performance of Speculative Decoding \citep{Leviathan2022FastIF}, not TTFT. Other methods explore using a larger teacher model to distill knowledge into a smaller student model to improve efficiency \citep{Bagherinezhad2018LabelRI,Gou2020KnowledgeDA,Xu2024ASO}. After training, the teacher model is discarded. Our method differs in that both models are retained.

\textbf{General Techniques for Efficiency:} Many works have explored quantization \citep{Lin2023AWQAW,Tseng2024QuIPEB, Frantar2022GPTQAP, Dettmers2023SpQRAS,Shao2023OmniQuantOC,Egiazarian2024ExtremeCO}, pruning \citep{AlizadehVahid2023LLMIA,Ma2023LLMPrunerOT,Zheng2024LearnTB}, and efficient design \citep{Mehta2024OpenELMAE,Zhang2024TinyLlamaAO,Abdin2024Phi3TR} to improve LLM efficiency. These techniques are orthogonal to ours, as they can be combined with our method or our baselines.

\section{Motivation: Time to First Token (TTFT)} \label{sec:motivation}
%Standard inference with an autoregressive transformer model begins with a prompt processing phase, in which the initial input of $N$ tokens is given to the model and the first output token is generated. Afterwards, the model operates in generation mode, in which tokens are autoregressively generated by feeding $1$ token to the model at a time. The time to first token (TTFT) is defined as how long a user must wait before the model produces the first token. TTFT directly translates to latency in the user experience.
%\qc{can we find/cite an example of user experience? like delay of the first turn of a chatbot conversation, or code completion, summarization etc.}
We observe that the TTFT of running a language model on an edge device can be prohibitively high, negatively impacting user experience. We visualize total prompt processing time of OpenELM \citep{Mehta2024OpenELMAE} models on an M2 Pro CPU \citep{everymac} with 32GB of RAM in \cref{fig:timing-cpu}. As prompt lengths and batch sizes increase, the TTFT transitions from noticeable to intolerable. Most users are not willing to wait more than a few seconds for a response from a chat bot \citep{chatbot}, but on-device prompt processing times can far exceed this. This long processing time would be further exacerbated by running the model on low-end hardware instead of a high-end laptop CPU.

In addition to the negative user experience created by high TTFTs, we note that the ratio of prompt processing time to generation time is of particular significance to applications that generate short responses. For example, in a question-answering system, the model may only generate a few tokens. Thus, the fractional runtime spent during prompt processing can dominate the overall runtime. In \cref{fig:timing-cpu}, we visualize the ratio of TTFT to the autoregressive generation time for an OpenELM 3B model. As prompt length and batch size increases, the process becomes compute-bound and the cost of prompt processing substantially increases relative to the cost of a generation step.

To illustrate the impact of this high ratio of prompt processing time to generation time, consider a question-answering system with $1024$ tokens of context, operating at batch size $1$, and generating a $2$ token response. In this case, the model will spend $4.8$ seconds processing the prompt and generating the first token, and only $4.8/12=0.4$ seconds generating the second token (since prompt processing takes $12$ times as long as generation, see \cref{fig:timing-cpu}). In this case, $5.2$ seconds are spent generating the output, of which $4.8/5.2=92.3\%$ is spent in prompt processing.

In summary, large prompt processing times can negatively impact practical usage of a model in $2$ ways. First, they can result in a large TTFT, negatively impacting user experience. Second, they also can represent a large fraction of the total compute used in inference. Our goal is to reduce on-device TTFT through a novel modeling strategy which we describe in the next section.

\section{KV Prediction} \label{sec:method}
% Inference with an autoregressive transformer proceeds in 2 phases. In the prompt processing phase, the $N$-token input query is given to the model, which (1) processes the prompt and builds the KV cache, and (2) outputs the first token generated by the model. After the prompt processing phase, the model operates in the generation phase, in which the model processes the previously output token to generate a new token and add a new entry to the KV cache.

% Similarly, our inference method proceeds in prompt processing, followed by generation. In contrast to standard prompt processing, our method performs prompt processing in two steps: (1) a smaller auxiliary model processes the $N$-token query and predicts the KV cache needed by a larger base model (\cref{fig:diagram}), then (2) a $1$-token generation step with the base model produces the first token (which is fast compared to the prompt processing step, since only $1$ token is processed). Afterwards, the model moves into generation phase, in which only the base model is used autoregressively (as in the standard inference case).

Recent works have shown that the KV cache can be compressed with little or no loss in accuracy at generation time \citep{Cai2024PyramidKVDK,Li2024SnapKVLK,Xiao2023EfficientSL,Ge2023ModelTY}. Thus, we hypothesize that the KV cache for a model can be approximated efficiently to reduce on-device TTFT. Our approximation uses an efficient learned model to predict the KV cache.

Our intuition is that an accurate KV cache prediction method should use a model of similar structure to the base model, so that the structure of the KV cache can be modeled more accurately. Thus, we use a smaller learned auxiliary transformer to process the prompt. Then, a set of learned linear projections is used to predict the base model's KV cache using the auxiliary model's KV cache as input. \cref{fig:diagram} shows an overview of our method.

%Our method improves TTFT by using a small auxiliary transformer model to process the prompt. After the prompt is processed, the KV cache of the base model is predicting 

%We develop a method capable of improving TTFT by processing the prompt with a small auxiliary model. The KV cache produced by the auxiliary model is used to predict the KV cache needed by the original (base) model. Once the predicted KV cache is obtained, the base model proceeds in generation mode.

In the following subsections, we present details of our method. In \cref{sec:predicting-kv-caches}, we give an overview of training and inference. In \cref{sec:arch-details}, we provide more architectural details of our KV Prediction models. In \cref{sec:loss-function}, we describe our loss function. Finally, in \cref{sec:timing}, we analyze the reduction in TTFT.

\subsection{Predicting KV Caches} \label{sec:predicting-kv-caches}

\begin{figure}
\begin{subfigure}{0.38\textwidth}
    \centering
    \includegraphics[width=\textwidth]{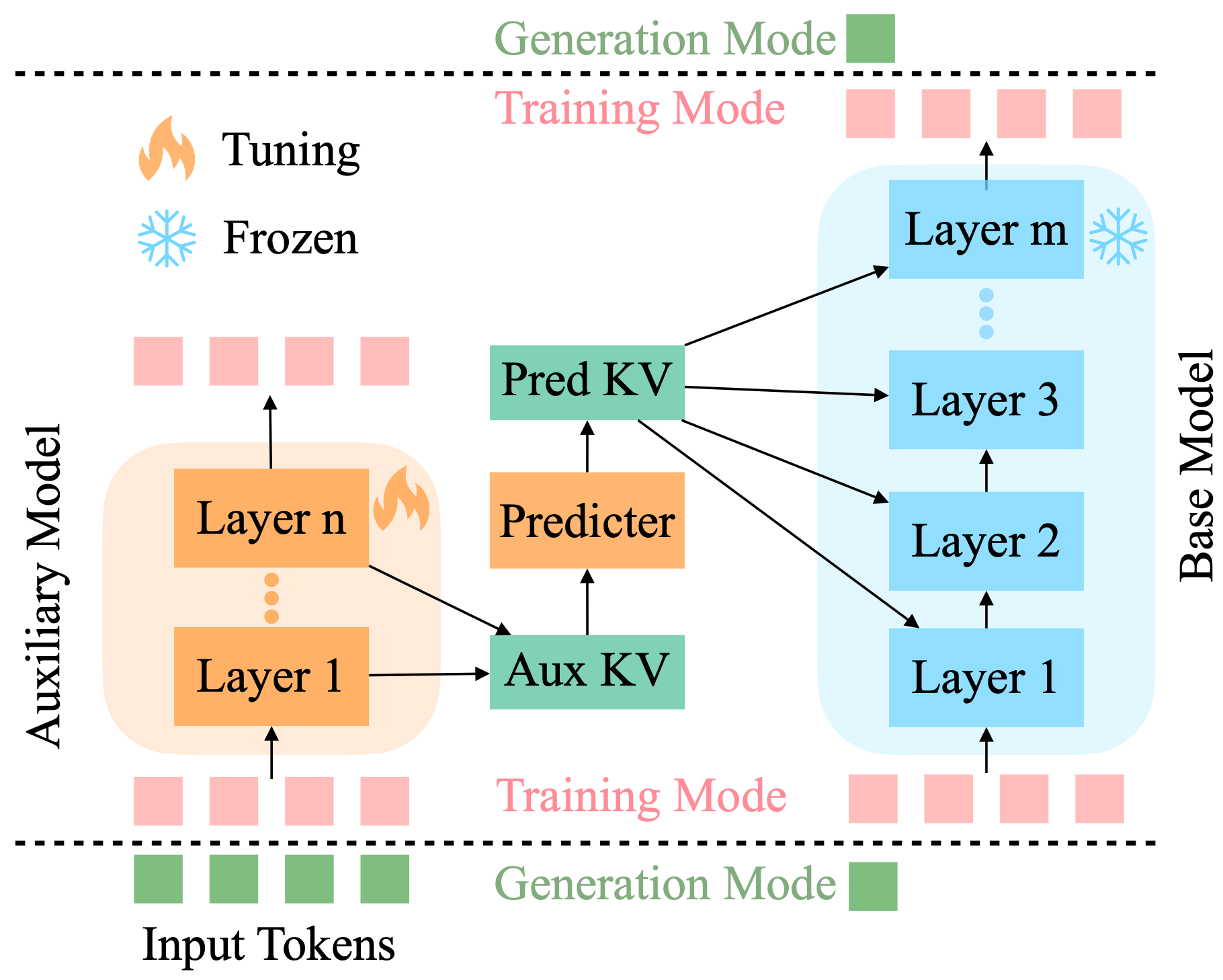}
    \caption{}
    \label{fig:diagram}
\end{subfigure}
\tiny
\begin{subfigure}{0.3\textwidth}
  \hrule
  \vspace{0.1cm}
  \textbf{KV Prediction Inference}
  \vspace{0.1cm}
  \hrule
  \vspace{0.1cm}
\begin{algorithmic}\vspace{0.05cm}
  \STATE {\textbf{Input:} base \base{}, aux \auxiliary{}, predictor \predictor{}, inputs \inputs{}}
    \STATE{\# Run \auxiliary{} to obtain logits and cache.}
  \STATE{$a, \auxiliarykv \leftarrow \auxiliary{}(\inputs{})$}
  \STATE{\# Predict the KV cache.}
  \STATE{$\predictedbasekv{} \leftarrow \predictor{}(\auxiliarykv{})$}
  \STATE{\# Run the base model on only the last token.}
  \STATE{$b \leftarrow \base{}(\inputs{}[-1], \predictedbasekv{})$}
  \STATE{\# Update inputs.}
  \STATE{$\inputs{} \leftarrow \inputs{} + \text{sample}(b)$}
  % \STATE{\# Generation loop.}
  \WHILE {generation not finished}
    \STATE{$b, p \leftarrow \base{}(\inputs{}[-1], \predictedbasekv{})$}
    \STATE{$\inputs{} \leftarrow \inputs{} + \text{argmax}(b)$}
    \STATE{\# Update KV cache.}
    \STATE{$\predictedbasekv{} \leftarrow \predictedbasekv{} + p$}
  \ENDWHILE
% \RETURN \inputs{}
\vspace{0.1cm}
\hrule

\end{algorithmic}

\subcaption{\label{alg:full_alg}}
\label{fig:algorithm-inference}
\end{subfigure}
\begin{subfigure}{0.3\textwidth}
  \hrule
  \vspace{0.1cm}
  \textbf{KV Prediction Training (One Iteration)}
  \vspace{0.1cm}
  \hrule
  \vspace{0.1cm}
\begin{algorithmic}
% \vspace{-0.1cm}
  \STATE {\textbf{Input:} base \base{}, aux \auxiliary{}, predictor \predictor{}, inputs \inputs{}, targets \targets{}, loss function $\mathcal{L}$} 
  \STATE{\# Run \auxiliary{} to obtain logits and KV cache.}
  \STATE{$a, \auxiliarykv \leftarrow \auxiliary{}(\inputs{})$}
  \STATE{\# Predict the KV cache.}
  \STATE{$\predictedbasekv{} \leftarrow \predictor{}(\auxiliarykv{})$}
  \STATE{\# Run the base model with the predicted KV cache.}
  \STATE{$b \leftarrow \base{}(\inputs{}, \predictedbasekv{})$}
  \STATE{\# Run the base model again to get the ground-truth KV cache.}
  \STATE{$\basekv{} \leftarrow \base{}(\inputs{})$}
  \STATE{\# Compute loss}
  \STATE{$l \leftarrow \mathcal{L}(b, a, \targets{}, \predictedbasekv{}, \basekv{})$}
  \STATE{\# Use $l$ to update \auxiliary{}, \predictor{}. Move to next iteration.}%\RETURN \base{}, trained \auxiliary{}, trained \predictor{}
  \vspace{0.1cm}
  \hrule
\end{algorithmic}
\subcaption{}
\label{fig:algorithm-training}
\end{subfigure}

\caption{(a) An overview of our training and inference method. (b) Our inference method. (c) Our training method.}
\end{figure}

%\qc{we need to describe the workflow/idea first regardless of training and inference. Something like: The idea is to run prompt processing through a smaller auxiliary model to generate KV caches in a fast way, then projected those KV caches and feed to the larger base model and generate the first token.}

% Paragraph explaining training.
Our model contains a frozen pretrained base network \base{}, a learned auxiliary network \auxiliary{}, and a learned KV predictor \predictor{}. The auxiliary and predictor networks are used to generate the predicted KV cache efficiently. Afterwards, inference proceeds with the base model.

% Paragraph explaining inference:
\textbf{Inference:} During inference (\cref{fig:algorithm-inference}) the prompt is passed to the auxiliary model and the auxiliary KV cache \auxiliarykv{} is computed. Then, the predicted base KV cache $\predictedbasekv{} = \predictor{}(\auxiliarykv{})$ is computed. At this point, \auxiliary{} and \predictor{} are no longer required to continue inference. To generate the first token, a single-token generation step of \base{} is run, but with \predictedbasekv{} being used as the keys and values in the attention operations (instead of using the keys and values from the base model's QKV projections). The logits produced from this generation step are used to produce the first output token. Now that the first token has been produced, generation continues autoregressively in the standard fashion, with new KV cache entries being added as new tokens are processed.

\textbf{Training:} During training (\cref{fig:algorithm-training}), a sequence of tokens \inputs{} are fed to the auxiliary model \auxiliary{} to produce output logits \logitsa{} and a KV cache \auxiliarykv{}. Then, the predicted KV cache $\predictedbasekv{}=\predictor{}(\auxiliarykv{})$ is computed. Finally, a forward pass of \base{} is computed using the predicted KV cache \predictedbasekv{} (instead of using the keys and values from the base model's QKV projections) to produce output logits \logitsb{}. In other words, the base model computes a forward pass using masked cross-attention with the predicted KV cache to produce output logits. Errors in the base model's logits backpropagate through the frozen base model and into \auxiliary{} and \predictor{}.

%\label{sec:architectural-details}
% Paragraph with architectural details.

% \qc{Another point is that the OpenELM example is scattered, we can either make it a full example after we explain the detail or make it more general framing and avoid the example specifics, otherwise it feels distracting. \cref{table:arch} needs to be general instead of OpenELM since our KV prediction method applies to any model. One table might be sufficient, IMO, we should make it more formal, no need to show all 32 layers.}

\subsection{Architectural Details} \label{sec:arch-details}
% \qc{this section needs to be more structured. we can say something like Key Designs in KV Prediction. And a few paragraphs, like Auxiliary Model Initialization, Predictor Network Architecutre, KV Cache Project. }
Our KV Prediction models are fully specified by our base, auxiliary, and predictor networks. Our base network always consists of a standard pretrained frozen transformer network. Here we describe the architectural details of our auxiliary and predictor networks. 

\textbf{Auxiliary Networks:} We use two different methods for choosing an auxiliary network. In the first method, referred to as \texttt{KVP-C} (KV Prediction with a Canonical model), we choose the auxiliary network to be a smaller model in the same family as the base model. In the second method for choosing an \texttt{auxiliary} network, referred to as \texttt{KVP-LP} (KV Prediction with a Layer-Pruned model), our auxiliary network consists of a copy of the base network with some of the layers removed.

\textbf{Predictor Networks:} For our \emph{predictor} network, we use a simple set of learned linear transforms to predict each layer of the base model's KV cache independently. We choose linear transforms primarily for their efficiency, as our focus is on reducing TTFT.

First, we need to choose which layer of the auxiliary cache to use to predict each layer of the base cache. To do this, we define a mapping from auxiliary cache layers $j$ to base cache layers $i$. Let $n_{\base{}}$ be the number of transformer layers in the base network, and $n_{\auxiliary{}}$ be the number of transformer layers in the auxiliary network.
 For each $i \in [0, ..., n_{\base{}}-1]$, let $j_i \in [0, ..., n_{\auxiliary{}}-1]$ denote the $j_i$th layer of the auxiliary cache $\auxiliarykv{}_{j_i}$. We will use $\auxiliarykv{}_{j_i}$ as input to a linear function to predict the layer of the base cache $\predictedbasekv{}_i$.
 
 Now that we have defined our mapping $j_i$, we define our set of linear functions. Our linear functions are defined as $\mathcal{F}_i(\auxiliarykv{}_{j_i}): \mathcal{R}^{d_{\auxiliarykv{}_{j_i}}} \to \mathcal{R}^{d_{\basekv{}_i}}$, where $d_{\auxiliarykv{}_{j_i}}$, $d_{\basekv{}_i}$ are the feature dimensions of the auxiliary and base KV caches at layers $j_i$ and $i$, respectively. We use these linear functions to compute each layer of the base KV cache, $\predictedbasekv{}_i=\mathcal{F}_i(\auxiliarykv{}_{j_i})$. Once each layer $\predictedbasekv{}_i$ is predicted, we concatenate them to produce $\predictedbasekv{}$.

%Let $n_B$ be the number of transformer layers in the base network, and $n_A$ be the number of transformer layers in the auxiliary network. For each layer $i \in [0, ..., n_B-1]$ in the predicted KV cache \predictedbasekv{}, the predictor has a corresponding linear function $\mathcal{F}_i(\auxiliarykv{}_j): \mathbb{R}^{d_{A,j_i}} \to \mathbb{R}^{d_{B,i}}$, where $j_i \in [0, ..., n_A-1]$ is the layer of \auxiliarykv{} used as input when predicting $\predictedbasekv{}_i$, and $d_{A, j_i}$, $d_{B, i}$ are the number of hidden dimensions at layer $i$ and $j$ of the base and auxiliary KV cache, respectively. Thus, each $\mathcal{F}_i$ independently projects tokens in layer $j$ of the auxiliary KV cache to the proper dimensionality for storage in layer $i$ of the base KV cache.

When choosing the auxiliary KV cache layer $j_i$ to use as the input to $\mathcal{F}_i$, we build on top of the observation in \cite{Brandon2024ReducingTK} that neighboring transformer layers will have KV caches that are more similar than distant layers. Thus, we use the first layer of \auxiliarykv{} as input to predict the first several layers of \basekv{}, and the second layer of \auxiliarykv{} as input to predict the next several layers of \basekv{}, and so forth. When $n_{\auxiliary{}}$ does not divide evenly into $n_{\base{}}$, we perform rounding.

\textbf{OpenELM KV Prediction Specification:}
When experimenting with KV Prediction, we use OpenELM \citep{Mehta2024OpenELMAE} models. We choose OpenELM because of its variable KV cache size in each layer, allowing us to evaluate our method in this challenging scenario.

For KVP-C experiments, we use smaller models from the OpenELM family as the auxiliary models. For KVP-LP experiments, we define a set of layer-pruned OpenELM models in \cref{sec:lp-archs}, which we use as the auxiliary model.

Our KV Prediction models are listed in \cref{table:arch}. Each model is specified by a base network, an auxiliary network, and the pattern of input auxiliary layers $j_i$ used to predict base KV cache layer $i$. When referring to a KV prediction architecture, we specify its base network, followed by either KVP-C or KVP-LP, followed by its auxiliary network (e.g. OE1.1B-KVP-C-450M).

\input{arch_table}

\subsection{Loss Function} 
\label{sec:loss-function}
\begin{table}
\centering

\begin{tabular}{c || c | c | c}
    \textbf{$\lambda_C$} & 0 & 1/28 & 1 \\
    \hline
    Acc & 8.95 & \textbf{17.38} & 15.96 \\
\end{tabular}

\caption{Ablation on TriviaQA (1-shot) of our choice of consistency loss coefficient $\lambda_C$. Our model is OpenELM1.1B-KVP-LP-0.50.\protect\footnotemark}
\label{table:ablation-loss-coefficient}
\end{table}

In this subsection we describe our training loss. It consists of three components: the base loss $\mathcal{L}_B$, the auxiliary loss $\mathcal{L}_A$, and the consistency loss $\mathcal{L}_C$.

\noindent\textbf{\emph{Base loss:}} The base loss $\baseloss{}=\mathcal{C}_B(\logitsb{}, \targets{}{})$ is the cross-entropy loss between the base model's outputs and the ground-truth labels $\targets{}{}$. The base model is frozen, but the gradients flow backwards through the KV predictor and the auxiliary model. This loss helps ensure that the predicted KV cache is compatible with the base model.

\footnotetext{Results differ slightly from other tables, as training hyperparameters were not finalized during this ablation.}

\noindent\textbf{\emph{Auxiliary loss:}} The auxiliary loss $\auxiliaryloss{}=\mathcal{C}_A(\logitsa{}, \targets{}{})$ is the cross-entropy loss between the auxiliary model's outputs and the ground-truth labels. Since the auxiliary logits \logitsa{} are never used during inference, this loss is not strictly required to produce an auxiliary model and a predictor model that can support KV prediction. However, in preliminary experiments we find that \auxiliaryloss{} improves training convergence slightly.

\noindent\textbf{\emph{Consistency loss:}} The consistency loss $\consistencyloss{}=\mathbb{L}_1(\predictedbasekv{}, \basekv{})$ is used to align the predicted KV caches with the base model's KV caches. The consistency loss is necessary to ensure that the predicted KV cache is compatible with the base model. To illustrate this point, we present an ablation in \cref{table:ablation-loss-coefficient} showing a model trained with the consistency loss coefficient set to $0$ ($\lambda_C=0$). Without the consistency loss, the predicted KV cache performs poorly.

%During generation, the auxiliary model is unused. Instead, the keys and values are computed by the base model, and added to the KV cache. Thus, it is important that the KV caches entries generated by the predictor (during prompt processing) are compatible with those generated by the base model (during generation). To ensure this, we enforce a direct loss term between the predicted caches \predictedbasekv{} and the target caches \basekv{}.

%In this evaluation, the predicted cache is used as the KV  However, on generative question answering (TQA column), performance drops sharply for $\lambda_C=0$. This is because generative question answering requires the KV cache produced by the base model to be aligned 

Our final loss is a simple weighted sum of the individual losses:

\begin{align}
    \mathcal{L(\logitsb{}, \logitsa{}, \targets{}, \predictedbasekv{}, \basekv{})} &= \lambda_B \baseloss{} + \lambda_A \auxiliaryloss{} + \lambda_C \consistencyloss{} \\
      &= \lambda_B \mathcal{C}_A(\logitsb{}, \targets{}{}) + \lambda_A \mathcal{C}_A(\logitsa{}, \targets{}{}) + \lambda_C \mathbb{L}_1(\predictedbasekv{}, \basekv{})
\end{align}

We found that the loss terms \baseloss{} and \auxiliaryloss{} were balanced by simply setting $\lambda_B=\lambda_A=1$. To choose $\lambda_C$, we perform an ablation between summing the loss across layers ($\lambda_C=1$) and averaging the loss across layers ($\lambda_C=1/n_B$) in \cref{table:ablation-loss-coefficient}. We found that $\lambda_C=1/n_B$ performed better.

\subsection{Improvements in TTFT FLOPs} \label{sec:timing}

% Paragraph mentioning that prompt processing time in theory should be N times as computationally expensive.
We analyze the improvement in TTFT of our method. The FLOPs-per-token compute cost of transformers inference can be estimated as $2P$, where $P$ is the number of parameters in the model \citep{Kaplan2020ScalingLF}. The total FLOPs required for prompt processing for $N$ tokens is $NP$. Thus, the ratio of the FLOPs of prompt processing to the ratio of FLOPs for a single generation step is $N$.% As seen in \cref{fig:timing-ratio-cpu}, the true ratio is below $N$, but prompt processing still often takes an order of magnitude longer than generation.

% Paragraph mentioning that on GPUs, compute is abundant. So, the bottleneck is memory not flops. But on edge devices, compute is less available.
%In practice, the ratio of time spent processing the prompt to the time spent generating a token will be lower than $N$ when memory transfer speeds are the bottleneck in performance. However, the impact of higher FLOP requirements for long contexts can significantly increase prompt processing time relative to generation time. This phenomenon is dependent on the prompt length, the batch size, and the compute fabric. Higher prompt lengths and batch sizes will bring a prompt processing operation closer to being compute-bound, as more FLOPs are required in the forward pass. Additionally, high-end GPUs have a large availability of compute, but lower-end devices have less powerful compute and can become compute-bound at prompt lengths where a high-end GPU might be memory-bound.

% Paragraph explaining the theoretical runtime speedup.
Let $t_\mathcal{N}(N)$ denote the forward pass FLOPs of network $\mathcal{N}$ with $N$ input tokens. As described in \cref{sec:method}, producing the first output token using KV Prediction requires an N-token forward pass of \auxiliary{}, followed by an N-token forward pass of \predictor{}, then a single-token forward pass of \base{} (to generate the first output token using the predicted KV cache). The FLOPs taken to process the prompt and generate the first token is $t_\mathcal{A}(N) + t_\mathcal{P}(N) + t_\mathcal{B}(1) = N t_\mathcal{A}(1) + t_\mathcal{P}(N) + t_\mathcal{B}(1)$. The computational cost of $t_\mathcal{P}(N)$ is negligible compared to $N t_\mathcal{A}(1)$, as it contains only a linear layer of smaller dimensionality than the transformer's FFN layers. For sufficient $N$, $N t_\mathcal{A}(1) \gg t_\mathcal{B}(1)$, and the FLOPs of a single inference are dominated by $N t_\mathcal{A}(1)$. Thus, the relative improvement in prompt processing FLOPs over the standard inference setup can be approximated as $t_\mathcal{A}(1)/t_\mathcal{B}(1)$.
% As described in \cite{Kaplan2020ScalingLF}, we know $t_\mathcal{N}(N) = N t_\mathcal{N}(1)$ when $\mathcal{N}$ is a Transformer. 

\section{Experimental Setup} \label{sec:experimental-setup}
We experiment with our KV Prediction method to evaluate the improvement in TTFT and the accuracy retention. Here we present details of our experimental setup.

\textbf{KV Prediction Models:} We experiment with KV Prediction using OpenELM \citep{Mehta2024OpenELMAE}, as (1) it provides a suitable efficient architecture for on-device execution, and (2) its layers have variable KV cache sizes, allowing us to test our method's robustness to this challenging scenario.

%As discussed in \cref{sec:predicting-kv-caches}, we have two methods for choosing auxiliary models: KVP-C and KVP-LP. We refer to our models by concatenating (1) the base model's name, then (2) the method for choosing the auxiliary model, then (3) the auxiliary model name. Thus, our models using a base of OpenELM1.1B are OE1.1B-KVP-C-450M, OE1.1B-KVP-C-270M, OE1.1B-KVP-LP-0.75, OE1.1B-KVP-LP-0.50, and OE1.1B-KVP-LP-0.25. Similarly, our models using a base of OpenELM3B are OE3B-KVP-C-1.1B, OE3B-KVP-C-450M, OE3B-KVP-C-270M, OE3B-KVP-LP-0.75, OE3B-KVP-LP-0.50, and OE3B-KVP-LP-0.25.

% We use two strategies for choosing our auxiliary model. The first is using a smaller model in the OpenELM family (KVP-C, described in \cref{sec:predicting-kv-caches}). We refer to these models as ``OE\$\{BASE\_SIZE\}-KVP-C-\$\{AUXILIARY\_SIZE\} (for example, OE1.1B-KVP-C-450M, OE1.1B-KVP-C-270M, OE3B-KVP-C-1.1B,  OE3B-KVP-C-450M, and OE3B-KVP-C-270M). The second strategy is using a layer-pruned OpenELM model (KVP-LP, described in \cref{sec:predicting-kv-caches}). We refer to these as OE\$\{BASE\_SIZE\}-KVP-LP-\$\{AUXILIARY\_FRAC\_KEPT\_LAYERS\} (for example, OE1.1B-KVP-LP-0.75, OE1.1B-KVP-LP-0.50 OE1.1B-KVP-LP-0.25, OE3B-KVP-LP-0.75, OE3B-KVP-LP-0.50, and OE3B-KVP-LP-0.25).

To train our models, we reload the base and auxiliary model weights from pretrained OpenELM models (in the case of KVP-LP experiments, we only reload the unpruned layers into the auxiliary model). We follow the training hyperparameters of OpenELM, training on RefinedWeb \citep{Penedo2023TheRD}, ArXiv \citep{clement2019arxiv}, and Wikipedia \citep{wikidump}. We shorten the training regime to $70\text{k}$ iterations on 64 H100 GPUs at a token length of $2048$ and a batch size of $16$, since we are reloading pretrained weights. For code completion experiments, our training setup is the same, but we instead train on The Stack \citep{Kocetkov2022TheS3} and divide the learning rate by $10$. A complete list of configs appears in our code release for reproducibility.

\textbf{Baselines:} Few works have explored methods for TTFT improvement applicable to on-device inference. Thus, our baselines follow \cite{Fu2024LazyLLMDT}. Our first set of baselines uses token pruning. Our \textbf{RP} baseline consists of randomly pruning tokens from the input, sweeping across token retention ratios of [0.25, 0.50, 0.75]. Our \textbf{SP} baseline consists of probing the first $25\%$ of network layers to obtain attention weights, then pruning tokens that have low attention scores and rerunning on the pruned tokens. We use token retention rates of [0.25, 0.50], omitting the higher retention rate of 0.75 since the overhead of processing $25\%$ of the query with the unpruned input means that lower retention ratios are needed to obtain a similar speedup to random pruning. Our \textbf{L} baseline consists of LazyLLM \citep{Fu2024LazyLLMDT}. LazyLLM prunes at progressively higher rates through the network, from an initial higher retention rate (or low pruning rate) to a final lower retention rate (or higher pruning rate). We adopt the standard configuration suggested by the authors of beginning pruning $30\%$ of the way into the network and ending pruning $90\%$ of the way into the network. For \textbf{L0.75}, our beginning retention rate (which is active $30\%$ of the way into the network) is $75\%$, and we sweep across ending retention rates of [0.75, 0.50, 0.25, 0.10, 0.05, 0.01]. For \textbf{L0.50}, our beginning retention rate is $0.50$, and we sweep across end retention rates [0.50, 0.25, 0.10, 0.05, 0.01].

Our second set of baselines sweeps across model sizes in the OpenELM family. Our \textbf{OE} baseline consists of OpenELM 1.1B, OpenELM450M, and OpenELM270M. Our \textbf{OE-LP} baseline consists of OpenELM layer-pruned models, fine-tuned with the same settings as our KV Prediction models. See \cref{sec:lp-archs} for a detailed description of these layer-pruned architectures.

%For our baselines, we use the pretrained models released in the official repository of OpenELM. We also experiment with using fine-tuned layer-pruned models to obtain baselines with the same TTFT as our KV Prediction models. We refer to these layer-pruned models (which don't use KV Prediction) as \$\{BASE\_MODEL\}-\$\{NUM\_LAYERS\} (for example, OE1.1B-0.25 corresponds to an OE1.1B model with $25\%$ of the layers retained and the rest removed).

%For OpenELM and TinyLlama baselines, we use publicly available pretrained models. For KV Prediction models, we reload the base and auxiliary models from pretrained checkpoints (in the case of layer-pruned models, we reload all the layers that weren't pruned). Our fine-tuning uses RefinedWeb \citep{Penedo2023TheRD}, ArXiv \citep{clement2019arxiv}, and Wikipedia \citep{wikidump}. We fine-tune for 

\begin{figure}

\centering
\includegraphics[width=\textwidth]{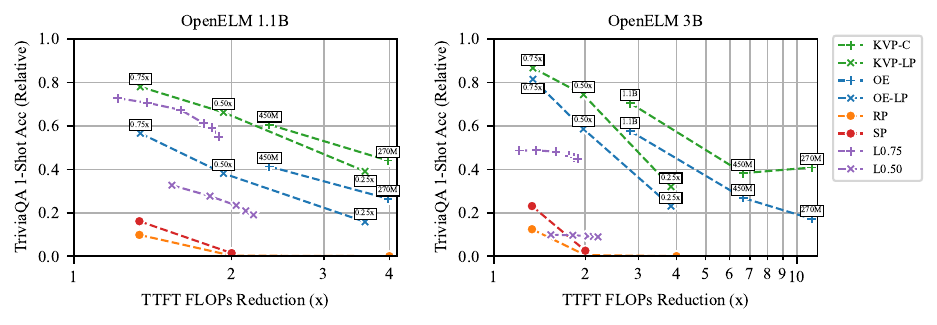}

  \caption{Efficiency-accuracy trade-off of our KV Prediction method (KVP-C, KVP-LP) on TriviaQA compared to baselines. The x-axis shows the relative reduction in FLOPs compared to the base network, and the y-axis shows the relative accuracy retention compared to the base network. (Left): Results using a base network of OpenELM 1.1B for KV Prediction. For KV Prediction models (green), points are annotated with the auxiliary network used. For example, the leftmost green ``x'' corresponds to OE1.1B-KVP-LP-0.75, and the leftmost green ``+'' corresponds to OE1.1B-KVP-C-450M. For OpenELM baselines (blue), points are annotated with the OpenELM variant used. All other baselines use variations of token pruning with different rates on OpenELM 1.1B (Right): Results using a base network of OpenELM 3B.}
  \label{fig:tqa}
\end{figure}
\section{Results} \label{sec:results}

We analyze our method's performance on language modeling benchmarks in the following sections. Our main results are generative evaluations, which allow us to assess the compatibility of the initial predicted KV cache with the autoregressively generated KV cache. In \cref{sec:tqa-results}, we investigate the improvement in TTFT of our method compared to baselines on question-answering with TriviaQA \citep{Joshi2017TriviaQAAL}, demonstrating that our method produces the pareto-optimal efficiency-accuracy tradeoff. In section \cref{sec:code-completion-results}, we demonstrate that our model also achieves the pareto-optimal efficiency-accuracy tradeoff on  python code completion with HumanEval \citep{Chen2021EvaluatingLL}, supporting the generality of our method. Finally, in \cref{sec:runtime-analysis}, we analyze the on-device runtime improvement in TTFT, demonstrating that our theoretical FLOPs reduction translates to runtime improvement on real hardware.

%Finally, in \cref{sec:kv-cache-consistency}, we analyze the compatibility of our predicted KV cache \predictedbasekv{} with the true KV cache \basekv{}.

\subsection{Question-Answering on TriviaQA} \label{sec:tqa-results}
We investigate our method's efficiency-accuracy trade-off for OpenELM 1.1B and OpenELM 3B in \cref{fig:tqa} (we also present these results in table format in \cref{sec:accuracy-values}). We measure accuracy on TriviaQA using LLMEvalHarness \citep{eval-harness}, and use the FLOPs speedup approximation developed in \cref{sec:timing}.

Our method produces the strongest efficiency-accuracy trade-off, tracing a pareto-optimal curve. At a fixed FLOPs reduction, our method achieves the highest accuracy. Equivalently, at fixed accuracy, our method obtains the highest FLOPs reduction. The KVP-C strategy outperforms the KVP-LP method, but the KVP-LP method is able to achieve higher accuracy retention because larger models can be obtained. For instance, OE3B-KVP-LP-0.75 (\cref{fig:tqa}, right side, top left point in the KVP-LP curve) has roughly $2\times$ as many parameters as OE3B-KVP-C-1.1B (\cref{fig:tqa}, right side, top left point in the KVP-C curve). In all cases except OE3B-KVP-LP-0.25, our model produces higher accuracy at equivalent TTFT FLOPs compared to baselines.

Our OE baselines correspond to only using the auxiliary model architecture from the KVP-C experiment (but with different weights). Directly comparing our KVP-C method to these baselines, we see that our method strongly increases accuracy. A similar observation can be made when comparing OE-LP and KVP-LP models.

Our naive token-pruning baselines of random pruning (RP) and static pruning (SP) do not perform very well. As explored in \cite{Fu2024LazyLLMDT}, SP can serve as a strong baseline when contexts are extremely long ($\sim4\text{k}$) tokens. We conjecture that performance is worse here because there are fewer redundant tokens in TriviaQA evaluations. LazyLLM provides much higher accuracy retention than naive token-pruning, but accuracy decreases as we push to more aggressive FLOP reductions.

\begin{figure}
\centering
\includegraphics[width=\textwidth]{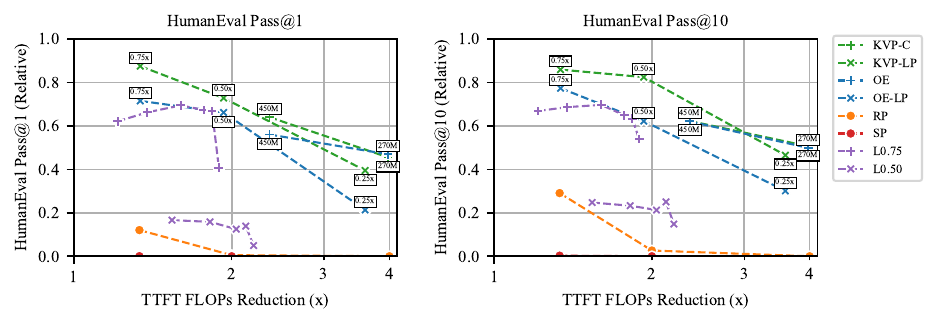}

  \caption{Efficiency-accuracy trade-off of our KV Prediction method (KVP-C, KVP-LP) compared to baselines on HumanEval python code completion. The x-axis shows the relative speedup in FLOPs compared to OpenELM 1.1B, and the y-axis shows the relative accuracy retention compared to OpenELM 1.1B. (Left): HumanEval Pass@1. (Right): HumanEval Pass@10 (Note that results for KVP-C and OE are overlapping.)}
  \label{fig:he}
\end{figure}
\subsection{Code Completion Results} \label{sec:code-completion-results}
We investigate our method's efficiency-accuracy trade-off for code completion with a base model of OpenELM 1.1B in \cref{fig:he} (we also present these results in table format in \cref{sec:accuracy-values}). We train our models on the Stack \citep{Kocetkov2022TheS3} and measure performance on HumanEval's python code completion benchmark \citep{Chen2021EvaluatingLL}.

We find that our model produces the strongest efficiency-accuracy trade-off, tracing a pareto-optimal curve.  As in the case of TriviaQA, our KVP-C strategy outperforms the KVP-LP method, but the KVP-LP method is able to achieve a higher accuracy retention because larger models can be chosen.

Our method obtains a much stronger efficiency-accuracy trade-off than OE and OE-LP baselines. Our KV prediction method also improves over token-pruning baselines (RP, SP, L0.75, and L0.50) in terms of efficiency and accuracy. The poor performance of token pruning methods can be attributed to the fact that code tokens have a tendency to carry less redundant information than general question-answering, making pruning more difficult.

\begin{figure}
    \begin{subfigure}{0.5\columnwidth}
    \centering
    \includegraphics[width=\textwidth]{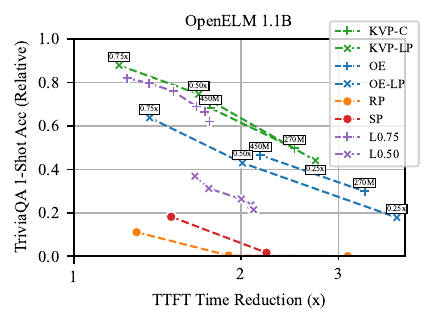}
        \vspace{-.75cm}
    \caption{}
    \label{fig:timing-tqa}
  \end{subfigure}
    \begin{subfigure}{0.5\columnwidth}
    \centering
    \includegraphics[width=\textwidth]{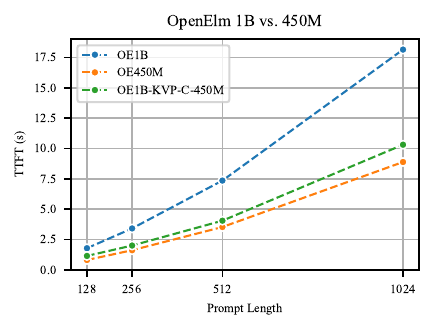}
    \vspace{-.75cm}
    \caption{}
    \label{fig:timing-oe1b-450m}
  \end{subfigure}
  \label{fig:timing}
  \caption{(\cref{fig:timing-tqa}): Accuracy on the TriviaQA dataset compared to benchmarked time to first token on CPU. (\cref{fig:timing-oe1b-450m}): The time to first token of our KVP prediction model OE1.1B-KVP-C-450M compared to OpenELM 1.1B and OpenELM450M.}
\end{figure}

\subsection{Timing Experiments} \label{sec:runtime-analysis}
We perform timing experiments to analyze the runtime improvement of our method. We measure the reduction in TTFT on an M2 Pro CPU \citep{everymac} with 32GB of RAM using the average query length of TriviaQA 1-shot (59 tokens) and a batch size of 64. We plot this TTFT reduction (relative to the Base model) and the accuracy retention (relative to the base model) in \cref{fig:timing-tqa}. We find that our method traces the pareto-optimal frontier. The TTFT of OE baseline models exceeds that of KVP-C models, but the stronger accuracy of the KVP-C models keeps them on the pareto-optimal frontier. A similar observation can be made for OE-LP and KVP-LP models. We also present these comparisons in table format in \cref{sec:accuracy-values}.

Next, we measure the TTFT of a KV Prediction model compared to its base and auxiliary models. To do this, we perform a comparison of the runtime characteristics of OpenELM 1.1B, OE1.1B-KVP-C-450M, and OpenELM 450M in \cref{fig:timing-oe1b-450m}. We set the batch size to 8 and show the change in TTFT as a function of prompt length. We find that the TTFT of OE1.1B-KVP-C-450M is within $10-20\%$ of the OpenELM 450M baseline, demonstrating that KV prediction has small overhead on-device relative to only running the auxiliary model. Both OE1.1B-KVP-C-450M and OE450M have a TTFT far superior to OpenELM 1.1B This indicates that our method is competitive with the auxiliary-only baseline in terms of speed, while being much more accurate (as discussed in \cref{sec:tqa-results}, \cref{sec:code-completion-results}).

\begin{figure}
\centering
  \begin{subfigure}{0.44\columnwidth}
    \centering
    \includegraphics[width=\columnwidth]{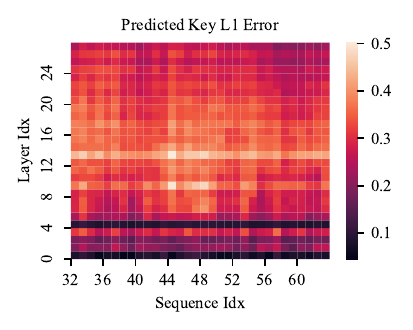}
        \vspace{-.7cm}
    \caption{}
    \label{fig:delta-keys}

    \end{subfigure}
    \begin{subfigure}{0.44\columnwidth}
    \includegraphics[width=\columnwidth]{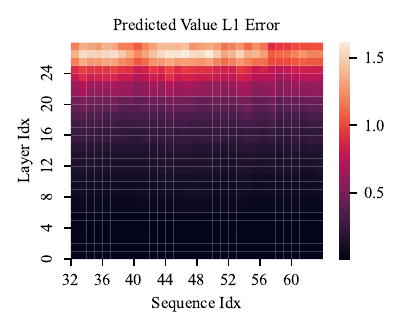}
    \vspace{-.7cm}
    \caption{}
    \label{fig:delta-values}

    \end{subfigure}
\vspace{-.4cm}
  \caption{L1 error of predicted keys (\cref{fig:delta-keys}) and values (\cref{fig:delta-values}) across the layer and sequence dimensions of OpenELM-1.1B-KVP-C-450M.}
  \label{fig:delta-layer-seq}
\end{figure}
\section{Analysis} \label{sec:analysis}

To better understand the performance of our KV cache prediction method, and to motivate future optimizations to improve performance, we analyze the quality of the KV cache predictions.

\textbf{L1 Error Across Sequences and Layers:}
We analyze the distribution of the L1 loss across layers and sequence indexes in \cref{fig:delta-layer-seq}. The errors are relatively stable across sequence index, with a few outliers. Across layers, the magnitude of key error is stable due to OpenELM's usage of key norm \citep{Henry2020QueryKeyNF}. Value error generally increases with depth due to propagation of errors.

\begin{table}
\centering
    \resizebox{0.8\textwidth}{!}{
\begin{tabular}{c | ccccccc | c | c}
    \textbf{Model} & arc-c & arc-e & boolq & hellaswag & piqa & sciq & winogrande & \textbf{Avg} & \light{TTFT Reduction} \\
    \midrule[1pt]
    OE 1.1B & 32.34 & 55.43 & 63.58 & 64.81 & 75.57 & 90.60 & 61.72 & 63.43 & \light{1.0} \\
    \midrule[1pt]
    OE1.1B-KVP-LP-0.75 & \textbf{30.97} & \textbf{54.42} & \textbf{59.79} & \textbf{62.91} & \textbf{74.92} & \textbf{90.40} & \textbf{62.04} & \textbf{62.21} & \light{1.34} \\
    OE 1.1B-0.75 & 29.52 & 51.81 & 57.52 & 58.38 & 73.88 & 87.70 & 60.46 & 59.90 & \light{1.34} \\
    \midrule[1pt]
    OE1.1B-KVP-LP-0.50 & \textbf{30.80} & \textbf{53.75} & 58.17 & \textbf{60.35} & \textbf{74.32} & \textbf{88.90} & \textbf{58.96} & \textbf{60.75} & \light{1.93} \\
    OE 1.1B-0.50 & 26.28 & 48.15 & \textbf{59.79} & 53.74 & 71.33 & 86.10 & 57.46 & 57.55 & \light{1.93} \\
    \midrule[1pt]
    OE1.1B-KVP-C-450M & \textbf{29.01} & \textbf{52.53} & \textbf{57.95} & \textbf{59.20} & \textbf{73.01} & \textbf{88.00} & \textbf{59.43} & \textbf{59.88} & \light{2.36} \\
    OE 450M & 27.56 & 48.06 & 55.78 & 53.97 & 72.31 & 87.20 & 58.01 & 57.56 & \light{2.36} \\
    \midrule[1pt]
    OE1.1B-KVP-LP-0.25 & \textbf{27.65} & \textbf{46.68} & 56.88 & \textbf{54.34} & \textbf{72.09} & \textbf{85.00} & \textbf{55.33} & \textbf{56.85} & \light{3.59} \\
    OE 1.1B-0.25 & 24.15 & 41.84 & \textbf{60.49} & 43.08 & 68.61 & 82.60 & 52.41 & 53.31 & \light{3.59} \\
    \midrule[1pt]
    OE1.1B-KVP-C-270M & \textbf{28.41} & \textbf{48.70} & 53.67 & \textbf{55.35} & \textbf{71.98} & \textbf{87.30} & \textbf{57.38} & \textbf{57.54} & \light{3.98} \\
    OE 270M & 26.45 & 45.08 & \textbf{53.98} & 46.71 & 69.75 & 84.70 & 53.91 & 54.37 & \light{3.98} \\
\end{tabular}
}

\caption{Comparison of OpenELM KV Prediction models with using only the auxiliary model or only the base model for Multiple-Choice Question Answering. We de-emphasize ``TTFT Reduction'' since the concept doesn't apply to multiple-choice question-answering evaluations.}
\label{table:mcqa-eval-1b}
\end{table}
\textbf{Multiple-Choice Question Answering:}
We analyze the quality of KV cache predictions by running multiple-choice question answering (MCQA) evaluations using the predicted cache as the keys and values for the base model. Since these MCQA evaluations don't produce output tokens, there is no notion of TTFT, and our method doesn't provide a speedup. The purpose of these evaluations is to measure the consistency of the predicted KV cache with the base KV cache through accuracy retention on MCQA.

In \cref{table:mcqa-eval-1b}, we present results for 7 MCQA tasks on OpenELM 1.1B and KV Prediction models. We directly compare each KV Prediction model to the results obtained by using only the auxiliary model. In all cases, the KV Prediction model obtains higher average accuracy. We also present MCQA evaluations for OpenELM 3B in \cref{sec:mcqa-3b}.

\textbf{Density of Differences:} In \cref{sec:density-of-differences}, we analyze the density of KV cache prediction errors. We find that the density of key errors is relatively consistent due to the fact that OpenELM uses key norm. However, the value errors increase in magnitude with network depth.

%Interestingly, the accuracy retention reates for our fastest KV Prediction models are significantly higher than for generative evaluation. This is likely because the base model is used to produce logits for all tokens (not just the last one), which may partially compensate for inaccuracies in KV predictions.

\section{Conclusion} \label{sec:conclusion}
We present a method for improving time to first token (TTFT) called KV Prediction. Our method uses a small auxiliary model to efficiently predict the KV cache needed by a larger base model. We analyze the theoretical and actual speedup of our model, as well as the accuracy retention. We find that our model maintains a strong efficiency-accuracy trade-off, creating a pareto-optimal trade-off in terms of accuracy retention and TTFT.

\ificlrfinal
\section{Acknowledgments}
We would like to thank Colorado Reed, Alvin Wan, Binazir Karimzadeh, Mengxia Yu, Christian Koguchi, Qi Shan, Rasoul Shafipour, Minsik Cho, Mahyar Najibi, Mehrdad Farajtabar and Qichen Fu for useful discussions and valuable feedback.
\else \fi

\bibliography{iclr2025_conference}
\bibliographystyle{iclr2025_conference}
\clearpage
\appendix
\begin{table}
    \centering
    \begin{subtable}{\textwidth}
        \centering
        \resizebox{\textwidth}{!}{
\begin{tabular}{c|c}
    \textbf{Model} & Layers Retained From OpenELM1.1B \\
    \hline
    OpenELM1.1B-0.75 & 0, 1, 2, 4, 5, 6, 8, 9, 10, 12, 13, 14, 16, 17, 18, 20, 21, 22, 24, 25, 26 \\
    \hline
    OpenELM1.1B-0.50 & 0, 2, 4, 6, 8, 10, 12, 14, 16, 18, 20, 22, 24, 26 \\
    \hline
    OpenELM1.1B-0.25 & 0, 4, 8, 12, 16, 20, 24
\end{tabular}
}
\caption{Specification of layer-pruned OpenELM 1.1B architectures.}
\label{table:lp-archs-1b}
\end{subtable}
    \begin{subtable}{\textwidth}
        \centering
        \resizebox{\textwidth}{!}{
\begin{tabular}{c|c}
    \textbf{Model} & Layers Retained From OpenELM3B \\
    \hline
    OpenELM3B-0.75 & 0, 1, 2, 4, 5, 6, 8, 9, 10, 12, 13, 14, 16, 17, 18, 20, 21, 22, 24, 25, 26, 28, 29, 30, 32, 33, 34 \\
    \hline
    OpenELM3B-0.50 & 0, 2, 4, 6, 8, 10, 12, 14, 16, 18, 20, 22, 24, 26, 28, 30, 32, 34 \\
    \hline
    OpenELM3B-0.25 & 0, 4, 8, 12, 16, 20, 24, 28, 32
\end{tabular}
}
\caption{Specification of layer-pruned OpenELM 3B architectures.}
\label{table:lp-archs-3b}
\end{subtable}
\caption{Specification of layer-pruned OpenELM architectures.}
\label{table:lp-archs}
\end{table}
\section{Layer-Pruned Architectures} \label{sec:lp-archs}
We provide details of our layer-pruned architectures here. Each architecture is defined by pruning layers at regular intervals from an existing OpenELM architecture. We specify the retained layers in \cref{table:lp-archs}.

\section{Accuracy Values} \label{sec:accuracy-values}
We give values used to produce plots. In \cref{table:oe1.1b-values}, we give accuracies and timing results for OpenELM 1.1B on TriviaQA.  In \cref{table:oe3b-values}, we give accuracies for OpenELM 3B on TriviaQA. In \cref{table:oe1.1b-values-he}, we give accuracies for OpenELM 1.1B on HumanEval.

\section{OpenELM 3B MCQA Evaluation} \label{sec:mcqa-3b}
In \cref{sec:analysis}, we presented multiple-choice question answering (MCQA) evaluations on KV Prediction models using an OpenELM 1.1B base. We present additional results on OpenELM 3B in \cref{table:mcqa-eval-3b}.

\section{Density of Differences in KV Prediction} \label{sec:density-of-differences}
\begin{figure}
    \centering
    \includegraphics[width=\textwidth]{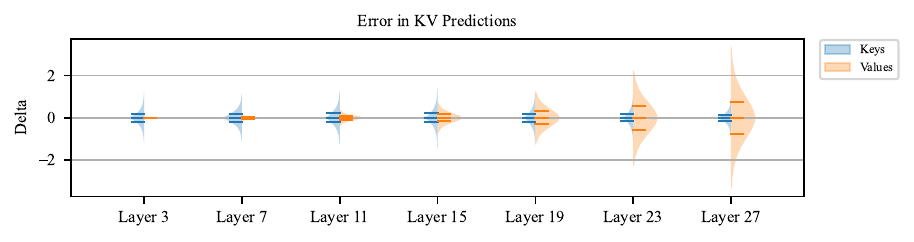}
    \vspace{-0.75cm}
  \caption{Distribution of the delta between KV cache predictions and targets in 7 different layers of OpenELM-1.1B-KVP-C-450M. The distributions of deltas for keys (blue) and values (orange) are shown separately.}
      \label{fig:error-density}
\end{figure}

In \cref{fig:error-density}, we analyze the distribution of the differences between the predicted KV cache and the target KV cache (e.g. similar to the \consistencyloss{} introduced in \cref{sec:loss-function}, but without the absolute value being computed). We pass a batch of data through our KV prediction model (to produce predictions \predictedbasekv{}) and through the base model (to produce targets \basekv{}), and compute the delta between the predictions and targets at every layer.

We observe that the delta for keys is stable, ranging roughly from -1 to 1 and centered at 0 at all layers. This is due to the fact that OpenELM uses normalized keys, so the distribution of the deltas is relatively consistent. By contrast, the error in predicted values increases with network depth. The distribution remains centered at 0, indicating that our cache prediction method is unbiased.
% Verdict: Flops (Rel), Time, Time (Rel), TQA, TQA (Rel) for 1.1B
% Abbreviated version fo 3B (no time).
% Separate graph with code completion.
\begin{table}
\centering
\resizebox{\textwidth}{!}{
\begin{tabular}{c|c|c|c|c|c}
\textbf{Model} & \textbf{FLOPs Reduction (Rel)} $\uparrow$ & \textbf{TTFT (s)} $\downarrow$ & \textbf{TTFT Reduction (Rel)} $\uparrow$ & \textbf{TQA} $\uparrow$ & \textbf{TQA (Rel)} $\uparrow$ \\
OE1.1B & 1.00 & 5.59 & 1.00 & 23.57 & 1.00 \\
\hline
OE450M & 2.36 & 2.58 & 2.17 & 10.97 & 0.41 \\
\hline
OE270M & 3.98 & 1.67 & 3.34 & 7.04 & 0.27 \\
\hline
OE1.1B-LP-0.75 & 1.34 & 4.08 & 1.37 & 15.04 & 0.57 \\
\hline
OE1.1B-LP-0.50 & 1.93 & 2.78 & 2.01 & 10.13 & 0.38 \\
\hline
OE1.1B-LP-0.25 & 3.60 & 1.46 & 3.82 & 4.21 & 0.16 \\
\hline
OE1.1B-KVP-C-450M & 2.36 & 3.17 & 1.76 & 16.09 & 0.61 \\
\hline
OE1.1B-KVP-C-270M & 3.98 & 2.24 & 2.50 & 11.71 & 0.44 \\ \hline
OE1.1B-KVP-LP-0.75 & 1.34 & 4.63 & 1.21 & 20.73 & 0.78 \\ \hline
OE1.1B-KVP-LP-0.50 & 1.93 & 3.33 & 1.68 & 17.59 & 0.66 \\ \hline
OE1.1B-KVP-LP-0.25 & 3.60 & 2.05 & 2.73 & 10.38 & 0.39 \\ \hline
RP-0.75 & 1.33 & 4.31 & 1.30 & 2.61 & 0.10 \\ \hline
RP-0.50 & 2.00 & 2.94 & 1.90 & 0.08 & 0.00 \\ \hline
RP-0.25 & 4.00 & 4.31 & 1.30 & 0.01 & 0.00 \\ \hline
SP-0.50 & 1.33 & 3.73 & 1.50 & 4.29 & 0.16 \\ \hline
SP-0.25 & 2.00 & 2.51 & 2.22 & 0.38 & 0.01 \\ \hline 
L0.75-0.75 & 1.21 & 4.47 & 1.25 & 19.33 & 0.73 \\ \hline
L0.75-0.50 & 1.38 & 4.09 & 1.37 & 18.78 & 0.71 \\ \hline
L0.75-0.25 & 1.60 & 3.69 & 1.51 & 17.89 & 0.67 \\ \hline
L0.75-0.10 & 1.89 & 3.18 & 1.76 & 14.63 & 0.55 \\ \hline
L0.75-0.05 & 1.83 & 3.24 & 1.72 & 15.70 & 0.59 \\ \hline
L0.75-0.01 & 1.77 & 3.36 & 1.66 & 16.25 & 0.61 \\ \hline
L0.50-0.50 & 1.54 & 3.38 & 1.65 & 8.69 & 0.33 \\ \hline
L0.50-0.25 & 1.82 & 3.19 & 1.75 & 7.36 & 0.28 \\ \hline
L0.50-0.10 & 2.04 & 2.79 & 2.00 & 6.21 & 0.23 \\ \hline
L0.50-0.05 & 2.13 & 2.68 & 2.09 & 5.56 & 0.21 \\ \hline
L0.50-0.01 & 2.20 & 2.65 & 2.11 & 5.06 & 0.19 \\
\end{tabular}
}
\caption{Relative FLOPs, runtime and accuracy values for OpenELM 1.1B on TriviaQA. Values are used to produce \cref{fig:tqa} (Left) and \cref{fig:timing-tqa}.}
\label{table:oe1.1b-values}
\end{table}

\begin{table}
\centering
\resizebox{\textwidth}{!}{
\begin{tabular}{c|c|c|c}
\textbf{Model} & \textbf{FLOPs Reduction (Rel)} $\uparrow$ & \textbf{TQA} $\uparrow$ & \textbf{TQA (Rel)} $\uparrow$ \\
OE3B & 1.00 & 40.87 & 1.00 \\ \hline
OE1.1B & 2.81 & 23.57 & 0.58 \\ \hline
OE450M & 6.64 & 10.97 & 0.27 \\ \hline
OE270M & 11.18 & 7.04 & 0.17 \\ \hline
OE3B-LP-0.75 & 1.34 & 33.31 & 0.81 \\ \hline
OE3B-LP-0.50 & 1.97 & 23.94 & 0.59 \\ \hline
OE3B-LP-0.25 & 3.84 & 9.43 & 0.23 \\ \hline
OE3B-KVP-C-1.1B & 2.81 & 28.83 & 0.71 \\ \hline
OE3B-KVP-C-450M & 6.64 & 15.64 & 0.38 \\ \hline
OE3B-KVP-C-270M & 11.18 & 16.70 & 0.41 \\ \hline
OE3B-KVP-LP-0.75 & 1.34 & 35.43 & 0.87 \\ \hline
OE3B-KVP-LP-0.50 & 1.97 & 30.41 & 0.74 \\ \hline
OE3B-KVP-LP-0.25 & 3.84 & 13.11 & 0.32 \\ \hline
RP 0.75 & 1.33 & 5.09 & 0.12 \\ \hline
RP 0.50 & 2.00 & 0.22 & 0.01 \\ \hline
RP0.25 & 4.00 & 0.02 & 0.00 \\ \hline
SP 0.50 & 1.33 & 9.45 & 0.23 \\ \hline
SP 0.25 & 2.00 & 1.04 & 0.03 \\ \hline
L0.75-0.75 & 1.21 & 19.90 & 0.49 \\ \hline
L0.75-0.50 & 1.38 & 19.93 & 0.49 \\ \hline
L0.75-0.25 & 1.60 & 19.68 & 0.48 \\ \hline
L0.75-0.10 & 1.89 & 18.35 & 0.45 \\ \hline
L0.75-0.05 & 1.83 & 18.58 & 0.45 \\ \hline
L0.75-0.01 & 1.77 & 18.95 & 0.46 \\ \hline
L0.50-0.50 & 1.54 & 4.04 & 0.10 \\ \hline
L0.50-0.25 & 1.82 & 3.95 & 0.10 \\ \hline
L0.50-0.10 & 2.04 & 3.81 & 0.09 \\ \hline
L0.50-0.05 & 2.13 & 3.72 & 0.09 \\ \hline
L0.50-0.01 & 2.20 & 3.66 & 0.09
\end{tabular}
}
\caption{Relative FLOPs and accuracy values of OpenELM 3B on TriviaQA. Values are used to produce \cref{fig:tqa} (Right).}
\label{table:oe3b-values}
\end{table}

\begin{table}
\centering
\resizebox{\textwidth}{!}{
\begin{tabular}{c|c|c|c|c|c}
\textbf{Model} & \textbf{FLOPs Reduction (Rel)} $\uparrow$ & \textbf{Pass@1} $\uparrow$ & \textbf{Pass@1 (Rel)} $\uparrow$ & \textbf{Pass@10} $\uparrow$ & \textbf{Pass@10 (Rel)} $\uparrow$ \\
OE1.1B & 1.00 & 15.79 & 1.00 & 23.07 & 1.00 \\ \hline
OE450M & 2.36 & 8.85 & 0.56 & 14.33 & 0.62 \\ \hline
OE270M & 3.98 & 7.41 & 0.47 & 11.51 & 0.50 \\ \hline
OE1.1B-LP-0.75 & 1.34 & 1.30 & 0.72 & 17.85 & 0.77 \\ \hline
OE1.1B-LP-0.50 & 1.93 & 10.46 & 0.66 & 14.36 & 0.62 \\ \hline
OE1.1B-LP-0.25 & 3.60 & 3.37 & 0.21 & 6.94 & 0.30 \\ \hline
OE1.1B-KVP-C-450M & 2.36 & 10.12 & 0.64 & 14.37 & 0.62 \\ \hline
OE1.1B-KVP-C-270M & 3.98 & 7.20 & 0.46 & 11.68 & 0.51 \\ \hline
OE1.1B-KVP-LP-0.75 & 1.34 & 13.82 & 0.88 & 19.83 & 0.86 \\ \hline
OE1.1B-KVP-LP-0.50 & 1.93 & 11.51 & 0.73 & 19.03 & 0.82 \\ \hline
OE1.1B-KVP-LP-0.25 & 3.60 & 6.24 & 0.39 & 10.73 & 0.47 \\ \hline
RP-0.75 & 1.33 & 1.90 & 0.12 & 6.71 & 0.29 \\ \hline
RP-0.50 & 2.00 & 0.08 & 0.01 & 0.61 & 0.03 \\ \hline
RP-0.25 & 4.00 & 0.00 & 0.00 & 0.00 & 0.00 \\ \hline
SP-0.50 & 1.33 & 0.01 & 0.00 & 0.06 & 0.00 \\ \hline
SP-0.25 & 2.00 & 0.00 & 0.00 & 0.00 & 0.00 \\ \hline
L0.75-0.75 & 1.21 & 9.81 & 0.62 & 15.43 & 0.67 \\ \hline
L0.75-0.50 & 1.38 & 10.46 & 0.66 & 15.85 & 0.69 \\ \hline
L0.75-0.25 & 1.60 & 10.97 & 0.69 & 16.08 & 0.70 \\ \hline
L0.75-0.10 & 1.77 & 10.63 & 0.67 & 15.00 & 0.65 \\ \hline
L0.75-0.05 & 1.83 & 10.56 & 0.67 & 14.62 & 0.63 \\ \hline
L0.75-0.01 & 1.89 & 6.44 & 0.41 & 12.46 & 0.54 \\ \hline
L0.50-0.50 & 1.54 & 2.63 & 0.17 & 5.71 & 0.25 \\ \hline
L0.50-0.25 & 1.82 & 2.50 & 0.16 & 5.37 & 0.23 \\ \hline
L0.50-0.10 & 2.04 & 1.97 & 0.12 & 4.91 & 0.21 \\ \hline
L0.50-0.05 & 2.13 & 2.19 & 0.14 & 5.77 & 0.25 \\ \hline
L0.50-0.01 & 2.20 & 0.77 & 0.05 & 3.42 & 0.15 \\
\end{tabular}
}
\caption{Relative FLOPs and accuracy values of OpenELM 1.1B on HumanEval. Values are used to produce \cref{fig:he}.}
\label{table:oe1.1b-values-he}
\end{table}

\begin{table}
\centering
    \resizebox{\textwidth}{!}{
    \begin{tabular}{c | ccccccc | c | c}
    \textbf{Model} & arc-c & arc-e & boolq & hellaswag & piqa & sciq & winogrande & \textbf{Avg} & \light{TTFT Reduction} \\
    \midrule[1.25pt]
    OE 3B & 35.58 & 59.89 & 67.40 & 72.44 & 78.24 & 92.70 & 65.51 & 67.39 & \light{1.0} \\
    \midrule[1pt]
    OE3B-KVP-LP-0.75 & \textbf{34.04} & 57.28 & \textbf{66.73} & \textbf{69.94} & \textbf{77.58} & \textbf{92.00} & \textbf{63.30} & \textbf{65.84} & \light{1.34} \\
    OE 3B-0.75 & 33.53 & \textbf{59.89} & 65.14 & 67.60 & 76.71 & \textbf{92.00} & 62.75 & 65.37 & \light{1.34} \\
    \midrule[1pt]
    OE3B-KVP-L-0.50 & \textbf{33.87} & \textbf{56.65} & \textbf{64.50} & \textbf{67.06} & \textbf{76.77} & \textbf{90.00} & 59.67 & \textbf{64.07} & \light{1.97} \\
    OE 3B-0.50 & 31.06 & 54.67 & 63.36 & 62.43 & 75.95 & 89.50 & \textbf{60.54} & 62.50 & \light{1.97} \\
    \midrule[1pt]
    OE3B-KVP-C-OE1.1B & \textbf{33.96} & \textbf{57.66} & \textbf{64.50} & \textbf{66.66} & \textbf{76.71} & 90.30 & 61.64 & \textbf{64.49} & \light{2.81} \\
    OE1.1B & 32.34 & 55.43 & 63.58 & 64.81 & 75.57 & \textbf{90.60} & \textbf{61.72} & 63.43 & \light{2.81} \\
    \midrule[1pt]
    OE3B-KVP-LP-0.25 & \textbf{29.27} & \textbf{51.39} & \textbf{62.63} & \textbf{58.87} & \textbf{74.59} & 85.30 & 56.27 & \textbf{59.76} & \light{3.84} \\
    OE3B-0.25 & 26.88 & 47.10 & 59.69 & 49.89 & 71.33 & \textbf{86.00} & \textbf{58.56} & 57.06 & \light{3.84} \\
    \midrule[1pt]
    OE3B-KVP-LP-450M & \textbf{30.97} & \textbf{54.00} & \textbf{62.84} & \textbf{61.94} & \textbf{74.76} & \textbf{88.50} & 57.46 & \textbf{61.50} & \light{6.64} \\
    OE450M & 27.56 & 48.06 & 55.78 & 53.97 & 72.31 & 87.20 & \textbf{58.01} & 57.56 & \light{6.64} \\
    \midrule[1pt]
    OE3B-KVP-C-270M & \textbf{29.27} & \textbf{49.87} & \textbf{59.48} & \textbf{59.09} & \textbf{73.50} & \textbf{87.50} & \textbf{56.35} & \textbf{59.30} & \light{11.18} \\
    OE 270M & 26.45 & 45.08 & 53.98 & 46.71 & 69.75 & 84.70 & 53.91 & 54.37 & \light{11.18} \\
\end{tabular}
}

\caption{Comparison of OpenELM 3B model variants on Multiple-Choice Question Answering. We de-emphasize ``TTFT Reduction'' since the concept of TTFT doesn't apply to multiple-choice question-answering evaluations.}
\label{table:mcqa-eval-3b}
\end{table}

\end{document}

%% file: math_commands.tex
\usepackage{amsmath,amsfonts,bm}

\def\eqref#1{equation~\ref{#1}}
\def\1{\bm{1}}

\DeclareMathAlphabet{\mathsfit}{\encodingdefault}{\sfdefault}{m}{sl}
\SetMathAlphabet{\mathsfit}{bold}{\encodingdefault}{\sfdefault}{bx}{n}

%% file: arch_table.tex
\begin{table}
    \begin{subtable}{\textwidth}
    \centering
    \resizebox{\textwidth}{!}{
    \begin{tabular}{c | c | c | c |c |c |c |c |c |c |c |c |c |c |c |c |c |c |c |c |c |c |c |c |c |c |c |c |c |c }
    \textbf{Model} & Aux & $j_{0}$ & $j_{1}$ & $j_{2}$ & $j_{3}$ & $j_{4}$ & $j_{5}$ & $j_{6}$ & $j_{7}$ & $j_{8}$ & $j_{9}$ & $j_{10}$ & $j_{11}$ & $j_{12}$ & $j_{13}$ & $j_{14}$ & $j_{15}$ & $j_{16}$ & $j_{17}$ & $j_{18}$ & $j_{19}$ & $j_{20}$ & $j_{21}$ & $j_{22}$ & $j_{23}$ & $j_{24}$ & $j_{25}$ & $j_{26}$ & $j_{27}$ \\
    \hline
    OE1.1B-KVP-C-450M & OE450M & 0 & 0 & 1 & 1 & 2 & 2 & 3 & 3 & 4 & 4 & 5 & 5 & 6 & 6 & 7 & 7 & 8 & 9 & 10 & 11 & 12 & 13 & 14 & 15 & 16 & 17 & 18 & 19 \\
    \hline
    OE1.1B-KVP-C-270M & OE270M & 0 & 0 & 1 & 1 & 2 & 2 & 3 & 3 & 4 & 4 & 5 & 5 & 6 & 6 & 7 & 7 & 8 & 8 & 9 & 9 & 10 & 10 & 11 & 11 & 12 & 13 & 14 & 15 \\
    \hline
    \hline
    OE1.1B-KVP-LP-0.75 & OE-LP-0.75 & 0 & 1 & 2 & 2 & 3 & 4 & 5 & 5 & 6 & 7 & 8 & 8 & 9 & 10 & 11 & 11 & 12 & 13 & 14 & 14 & 15 & 16 & 17 & 17 & 18 & 19 & 20 & 20 \\
    \hline
    OE1.1B-KVP-LP-0.50 & OE-LP-0.50 & 0 & 0 & 1 & 1 & 2 & 2 & 3 & 3 & 4 & 4 & 5 & 5 & 6 & 6 & 7 & 7 & 8 & 8 & 9 & 9 & 10 & 10 & 11 & 11 & 12 & 12 & 13 & 13 \\
    \hline
    OE1.1B-KVP-LP-0.25 & OE-LP-0.25 & 0 & 0 & 0 & 0 & 1 & 1 & 1 & 1 & 2 & 2 & 2 & 2 & 3 & 3 & 3 & 3 & 4 & 4 & 4 & 4 & 5 & 5 & 5 & 5 & 6 & 6 & 6 & 6 \\
    \end{tabular}
    }
    \caption{KV Prediction models using a base model of OpenELM 1.1B.}
    \label{table:arch-1b}
    \end{subtable}

    \begin{subtable}{\textwidth}
    \centering
    \resizebox{\textwidth}{!}{
    \begin{tabular}{c | c | c |c |c |c |c |c |c |c |c |c |c |c |c |c |c |c |c |c |c |c |c |c |c |c |c |c |c |c |c |c |c |c |c |c |c |c }
    \textbf{Model} & Aux & $j_{0}$ & $j_{1}$ & $j_{2}$ & $j_{3}$ & $j_{4}$ & $j_{5}$ & $j_{6}$ & $j_{7}$ & $j_{8}$ & $j_{9}$ & $j_{10}$ & $j_{11}$ & $j_{12}$ & $j_{13}$ & $j_{14}$ & $j_{15}$ & $j_{16}$ & $j_{17}$ & $j_{18}$ & $j_{19}$ & $j_{20}$ & $j_{21}$ & $j_{22}$ & $j_{23}$ & $j_{24}$ & $j_{25}$ & $j_{26}$ & $j_{27}$ & $j_{28}$ & $j_{29}$ & $j_{30}$ & $j_{31}$ & $j_{32}$ & $j_{33}$ & $j_{34}$ & $j_{35}$ \\ \hline
    OE3B-KVP-C-1.1B & OE1.1B & 0 & 0 & 1 & 1 & 2 & 2 & 3 & 3 & 4 & 4 & 5 & 5 & 6 & 6 & 7 & 7 & 8 & 9 & 10 & 11 & 12 & 13 & 14 & 15 & 16 & 17 & 18 & 19 & 20 & 21 & 22 & 23 & 24 & 25 & 26 & 27 \\ \hline
    OE3B-KVP-C-450M & OE450M & 0 & 0 & 1 & 1 & 2 & 2 & 3 & 3 & 4 & 4 & 5 & 5 & 6 & 6 & 7 & 7 & 8 & 8 & 9 & 9 & 10 & 10 & 11 & 11 & 12 & 12 & 13 & 13 & 14 & 14 & 15 & 15 & 16 & 17 & 18 & 19 \\ \hline
    OE3B-KVP-C-270M & OE270M & 0 & 0 & 0 & 1 & 1 & 1 & 2 & 2 & 2 & 3 & 3 & 3 & 4 & 4 & 5 & 5 & 6 & 6 & 7 & 7 & 8 & 8 & 9 & 9 & 10 & 10 & 11 & 11 & 12 & 12 & 13 & 13 & 14 & 14 & 15 & 15 \\
    \hline
    \hline
    OE3B-KVP-LP-0.75 & OE-LP-0.75 & 0 & 1 & 2 & 2 & 3 & 4 & 5 & 5 & 6 & 7 & 8 & 8 & 9 & 10 & 11 & 11 & 12 & 13 & 14 & 14 & 15 & 16 & 17 & 17 & 18 & 19 & 20 & 20 & 21 & 22 & 23 & 23 & 24 & 25 & 26 & 26 \\ \hline
    OE3B-KVP-LP-0.50 & OE-LP-0.50 & 0 & 0 & 1 & 1 & 2 & 2 & 3 & 3 & 4 & 4 & 5 & 5 & 6 & 6 & 7 & 7 & 8 & 8 & 9 & 9 & 10 & 10 & 11 & 11 & 12 & 12 & 13 & 13 & 14 & 14 & 15 & 15 & 16 & 16 & 17 & 17 \\ \hline
    OE3B-KVP-LP-0.25 & OE-LP-0.25 & 0 & 0 & 0 & 0 & 1 & 1 & 1 & 1 & 2 & 2 & 2 & 2 & 3 & 3 & 3 & 3 & 4 & 4 & 4 & 4 & 5 & 5 & 5 & 5 & 6 & 6 & 6 & 6 & 7 & 7 & 7 & 7 & 8 & 8 & 8 & 8 \\
    \end{tabular}
    }
    \caption{KV Prediction models using a base model of OpenELM 3B.}
    \label{table:arch-3b}
    \end{subtable}
    \caption{KV Prediction models. Each model is specified by a base network, an auxiliary network, and the mapping of input auxiliary layers $j_i$ to base layers $i$. (\cref{table:arch-1b}): models using a base network of OpenELM 1.1B. (\cref{table:arch-3b}): models using a base network of OpenELM 3B.}
    \label{table:arch}

\end{table}